\documentclass[10pt,twocolumn,letterpaper]{article}

\usepackage[pagenumbers]{cvpr}

\usepackage{twemojis}
\usepackage{booktabs}
\usepackage{xcolor}
\usepackage{pifont}
\newcommand{\cmark}{\textcolor{green!50!black}{\ding{51}}}
\newcommand{\xmark}{\textcolor{red}{\ding{55}}}
\usepackage{wasysym}

\newcommand*{\affaddr}[1]{#1}
\newcommand*{\affmark}[1][*]{\textsuperscript{#1}}

\usepackage{subcaption}
\usepackage{tikz}
\usetikzlibrary{matrix,calc}
\usepackage{microtype}

\newcommand{\PillOuterFrac}{0.98}
\newdimen\PillXPad   \PillXPad=4.0pt
\newdimen\PillYPad   \PillYPad=1.4pt
\newdimen\PillRule   \PillRule=0.25pt
\newdimen\PillRadius \PillRadius=3.0pt

\tikzset{
  cvprPill/.style={
    rounded corners=\the\PillRadius,
    draw=black!20,
    fill=black!2,
    line width=\the\PillRule,
    inner xsep=\the\PillXPad,
    inner ysep=\the\PillYPad,
    outer sep=0pt,
    line cap=round,
    line join=round
  }
}

\newsavebox{\capline}

\DeclareCaptionFormat{cvprpill}{%
  \begingroup
  \sbox{\capline}{\strut #1\hspace{0.35em}#3}%

  \dimen0=\dimexpr\PillOuterFrac\hsize-\PillXPad-\PillXPad-\PillRule-\PillRule\relax

  \noindent\hbox to\hsize{\hss
    \tikz[baseline=(cap.base)]\node[cvprPill] (cap) {%
      \microtypesetup{protrusion=false}%
      \ifdim\wd\capline<\dimen0
        \strut\noindent\ignorespaces#1\hspace{0.35em}#3\unskip
      \else
        \begin{minipage}{\dimen0}
          \microtypesetup{protrusion=false}%
          \setlength{\parindent}{0pt}\setlength{\parskip}{0pt}%
          \leftskip=0pt \rightskip=0pt
          \parfillskip=0pt plus 1fil
          \spaceskip=0pt \xspaceskip=0pt
          \hyphenpenalty=10000 \exhyphenpenalty=10000
          \tolerance=2000 \emergencystretch=1em
          \noindent\ignorespaces#1\hspace{0.35em}#3\unskip
        \end{minipage}
      \fi
    };%
  \hss}%
  \endgroup
}

\definecolor{cvprblue}{rgb}{0.21,0.49,0.74}
\usepackage[pagebackref,breaklinks,colorlinks,allcolors=cvprblue]{hyperref}

\def\paperID{}
\def\confName{CVPR}
\def\confYear{2026}

\title{BrickNet: Graph-Backed Generative Brick Assembly}

\author{
Peter~Kulits\affmark[1,2] \quad
Cordelia~Schmid\affmark[1]\\
{\small \affaddr{\affmark[1]Inria, \`Ecole Normale Sup\`erieure, CNRS, PSL Research University} \quad \affaddr{\affmark[2]Max Planck Institute for Intelligent Systems, T{\"u}bingen}}\\
\vspace{-1.0cm}
\url{https://kulits.github.io/BrickNet}
}

\begin{document}
\newcommand{\teaserCaption}{
We finetune an LLM to autoregressively generate LEGO-brick build sequences.
To enable this, we introduce a large-scale dataset of LDraw~\cite{ldraw} structures composed of parts (samples above), as well as a novel graph-backed parametrization to represent arbitrary objects.
}
\twocolumn[{
    \renewcommand\twocolumn[1][]{#1}
    \maketitle
    \centering
    \begin{minipage}{\linewidth}
        \centering
        \includegraphics[width=\linewidth]{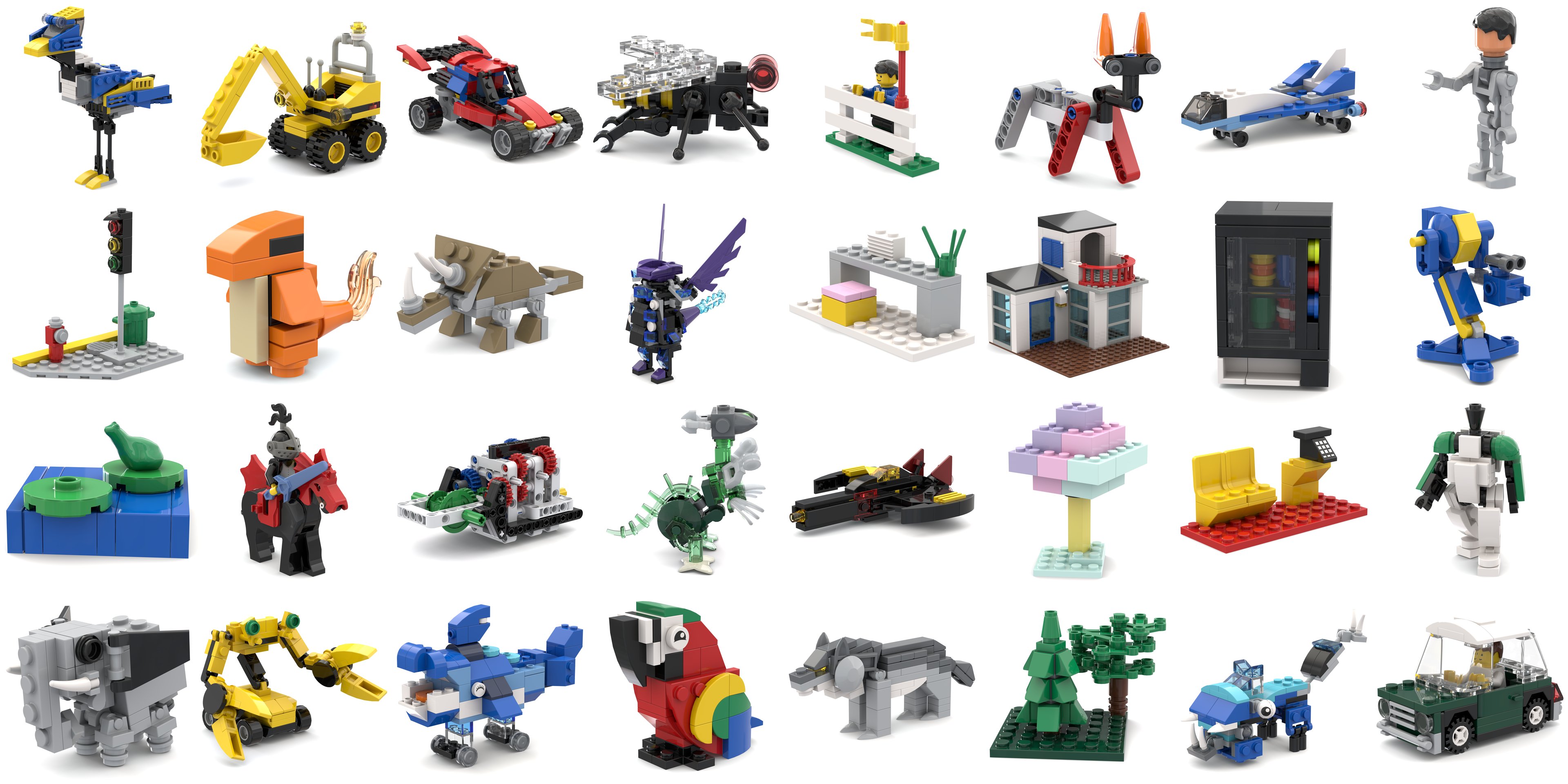}
    \end{minipage}
    {\captionsetup{hypcap=false}\captionof{figure}{\teaserCaption}\label{fig:teaser}}
    \vspace{0.6cm}
}]

\begin{abstract}
We train a language model to generate LEGO\textsuperscript{\tiny\textregistered}-brick build sequences.
While prior work has been restricted to discrete, voxel-like towers, we consider a much broader set of pieces, encompassing thousands of part types with diverse connection semantics.
To enable this, we first collect a large-scale dataset of over 100,000 human-designed LDraw brick objects and scenes.
The complexity of our setting makes it challenging to autoregressively assemble structures that satisfy physical constraints.
When predicting block pose directly, build sequences quickly become invalid after a small number of steps.
Although pieces are placed in 3D space, it is the spatial relationships of the parts which define the whole.
With this in mind, we design a graph-based program representation that parametrizes structure through connectivity, improving the physical grounding of generated sequences.
To enable future applications, we make our dataset and models available for research purposes.
\end{abstract}
\vspace*{-12pt}
    
\section{Introduction}\label{sec:introduction}
Many objects are naturally described at the level of parts and their configuration.
Recent work in 3D generation has explored representations that encode these relationships explicitly, including part graphs~\cite{structurenet} and executable shape programs~\cite{shapeassembly,csgnet}.
The resulting generative problem is then to produce objects whose global geometry and compositional structure are consistent.

Sequential LEGO assembly offers a compact instance of this broader problem.
A brick structure is defined not only by the arrangement of its parts but also by its construction.
Each part added must satisfy discrete rules, which constrain the set of valid continuations.
At the same time, the domain is also quite expressive, encompassing thousands of part types with rich connection diversity and semantics (\cref{fig:teaser}).

\begin{figure}[t]
\centering

\begin{subfigure}[b]{\linewidth}
\centering
\vspace{-0.25cm}
\includegraphics[width=\linewidth]{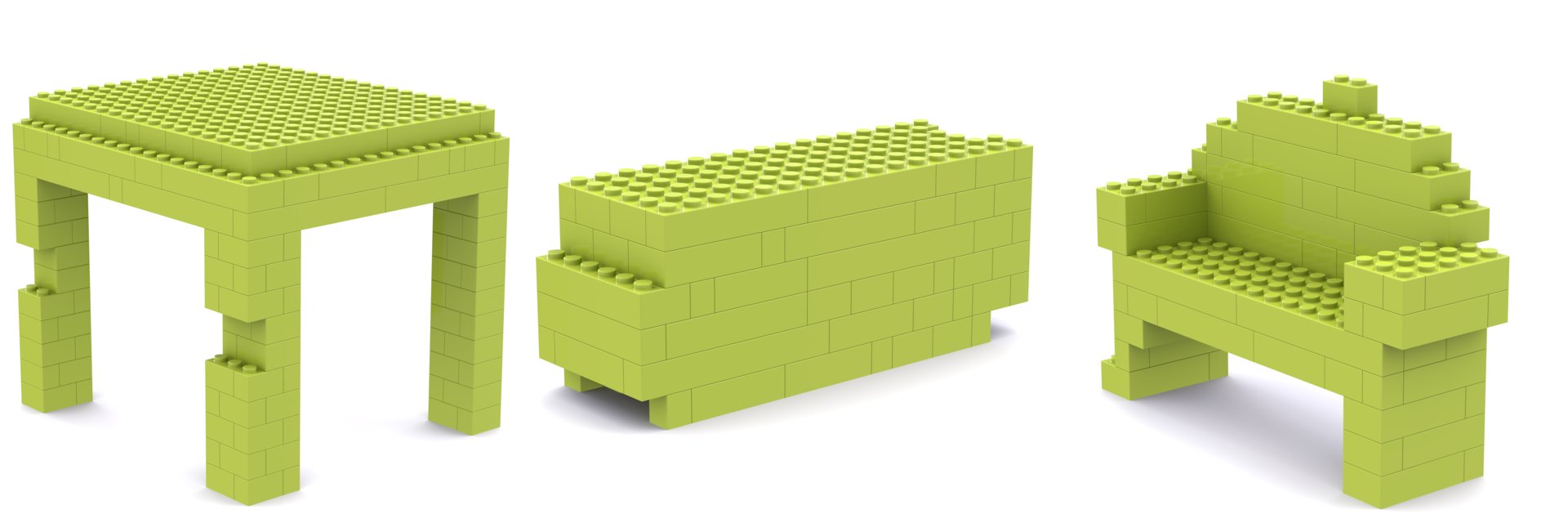}
\par\vspace{-0.3cm}
\caption{\textbf{BrickGPT~\citep{Pun_2025_ICCV} Samples}}
\label{fig:brickgpt_samples}
\end{subfigure}
\par\vspace{0.2cm}
\begin{subfigure}[b]{\linewidth}
\centering
\begin{minipage}[b]{0.40\linewidth}
\centering
\includegraphics[width=\linewidth]{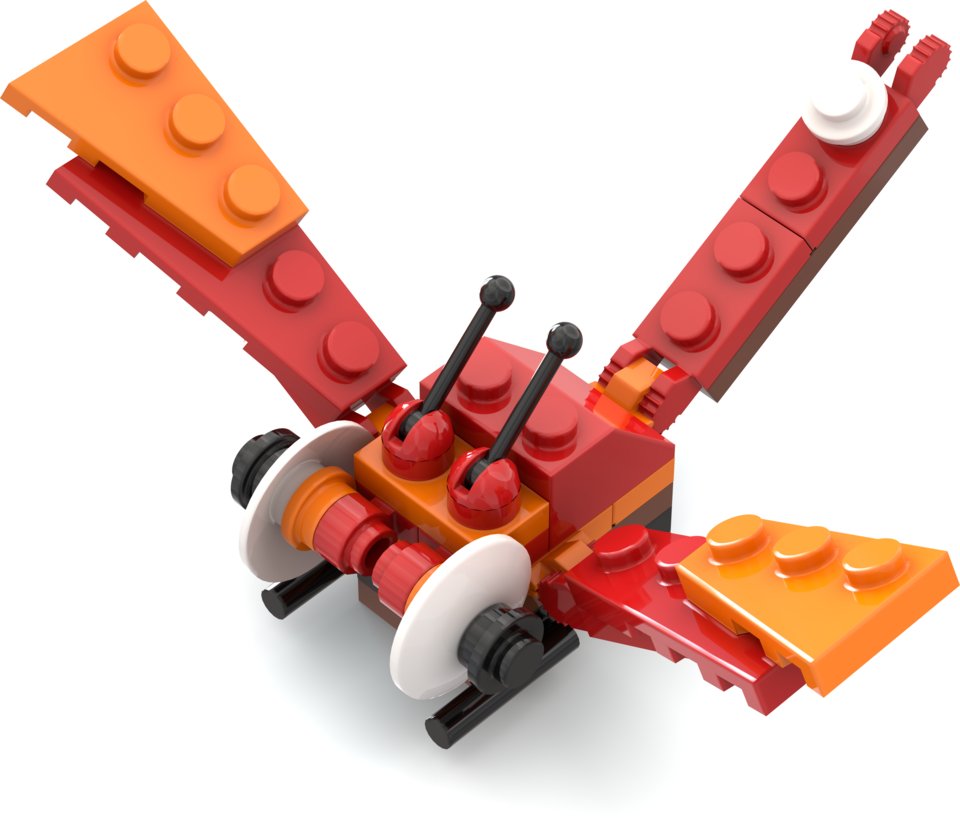}
\end{minipage}\hfill
\begin{minipage}[b]{0.58\linewidth}
\centering
\begin{tikzpicture}
\node[inner sep=0, anchor=south west] (img) at (0,0)
{\includegraphics[width=\linewidth]{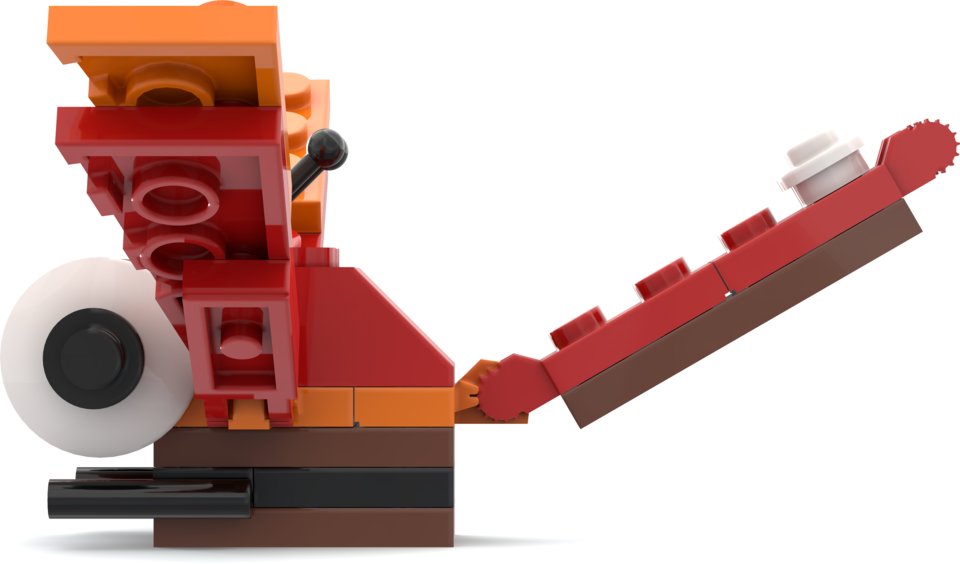}};
\begin{scope}[x={(img.south east)}, y={(img.north west)}]
\coordinate (oneStart) at (0.53,0.48);
\def\oneAngle{-110}
\def\oneLength{0.125}
\coordinate (oneTip) at ($(oneStart)+(\oneAngle:\oneLength)$);
\coordinate (fiveStart) at (0.75,0.79);
\def\fiveAngle{-27}
\def\fiveLength{0.089}
\coordinate (fiveTip) at ($(fiveStart)+(\fiveAngle:\fiveLength)$);
\newcommand{\dragonflyLabelOffset}{0.35}
\draw[->, line width=0.6pt] (oneStart) -- (oneTip)
node[pos=-\dragonflyLabelOffset, anchor=center, inner sep=0.5pt, font=\small] {1};
\draw[->, line width=0.6pt] (fiveStart) -- (fiveTip)
node[pos=-\dragonflyLabelOffset, anchor=center, inner sep=0.5pt, font=\small] {5};
\end{scope}
\end{tikzpicture}
\end{minipage}
\caption{\textbf{Human-Designed Sample}}
\vspace{-0.2cm}
\label{fig:real-world}
\end{subfigure}

\caption{\textbf{Motivation.}
To teach a model to autoregressively generate brick structures in a discrete, voxelized domain, it is intuitive to train it to regress 3D coordinates (\cref{fig:brickgpt_samples}).
However, doing so becomes more difficult when dealing with the complexities of real-world objects (\cref{fig:real-world}).
Starting at the orange hinge plate (1) and placing bricks down to the white stud (5) at the end requires maintaining a high degree of numerical precision across steps.
}
\label{fig:dragonfly}
\end{figure}

However, while a number of approaches for generative brick assembly have been proposed, they have been restricted to toy subsets of this representational space.
Typically assuming a discrete grid and only a handful of part types, triangular meshes can be voxelized to produce training samples (\cref{fig:brickgpt_samples}).
While useful as initial demonstrations, we argue such domains lack the expressivity which makes LEGO a particularly compelling domain for sequential generation.

One reason that this richer generation setting has not been studied is the lack of suitable training data, which can not as easily be bootstrapped.
To address this, we propose BrickNet, the first large-scale dataset of human-designed brick structures.
Curated from publicly available online sources, our dataset encompasses 320,808 samples, 9,743 part variants, and cumulatively 40,549,969 placed bricks.

Yet, from the increased expressivity arise representational challenges.
When restricted to a simple grid, up is up, the rotational frame is constant, and predicting coordinates seems natural.
However, real-world samples do not adhere to these assumptions.
Take, for example, the dragonfly in \cref{fig:real-world}.
To autoregressively assemble it from, say, the parts from 1 to 5, a model must keep track of the 6-DoF pose of each block in-between, and accumulate transforms along a shifted rotational frame, which becomes a problem of precision.

Instead, our insight is to make connectivity first-class.
To do so, we annotate each part with typed connectors and pairing semantics.
Then, we design a compact parametrization of structure that represents spatial relationships between parts through a graph of their connectivity.
In doing so, arbitrary structures can be serialized as a spanning tree, which can be executed to recover the original spatial transforms.

In summary, our primary contributions are:
\begin{enumerate}
\item A large-scale human-designed LDraw-structure dataset
\item Typed connectivity annotations for each of the parts
\item A graph-backed parametrization of part connectivity
\item Autoregressive models trained on our representation
\end{enumerate}

\section{Related Work}\label{sec:related}
\noindent\textbf{Bricks and Assembly.}
LEGO-brick assembly has been explored as both a generative and a reconstruction task.
Existing generative approaches have been restricted to discrete grids with small part vocabularies.
Some place bricks sequentially~\citep{combinatorial,blocks_assemble}, while others first predict a full volume and then decompose it into valid placements~\citep{brecs,micro_buildings}.
Similarly to our work, \citet{PEYSAKHOV_REGLI_2003} employ a graph parametrization of brick structure, but model only snap-fit connections between rectangular blocks and evolve structures using genetic algorithms rather that learning composition.
\citet{thompson} adopt this representation to train an autoregressive graph neural network, but instead are restricted to a single brick type and grid-aligned output positions.
BrickGPT~\citep{Pun_2025_ICCV} serializes block placements as text to finetune a language model to autoregressively produce build sequences as we do, but predict voxel positions.

For reconstruction, \citet{chung} train a reinforcement learning agent to assemble structures given multi-view images.
\citet{treesba} model assembly order as a breadth-first tree and train a model in a self-supervised manner.
\citet{mepnet} address a different input modality, inferring 3D block pose from visual instruction manuals by learning 2D--3D correspondence.
\citet{break_and_make,learning_to_buic} task an agent in a simulator to disassemble and reassemble structures.

\noindent\textbf{Program Synthesis for 3D Structure.}
A parallel line of work models 3D structure through executable programs rather than raw geometry.
CSGNet~\citep{csgnet} and InverseCSG~\citep{inversecsg} recover constructive solid geometry trees from images or 3D input.
ShapeAssembly~\citep{shapeassembly} generates hierarchical part-placement programs that can be executed to produce 3D shapes.
StructureNet~\citep{structurenet} models part-compositional structure through graphs of hierarchical relationships.
In a similar vein, we serialize brick structure as a spanning tree over a typed connectivity graph, where each edge is an instruction that can be realized to produce a local SE(3) transformation between parts.

\noindent\textbf{LLMs and 3D Understanding.}
LLMs have recently demonstrated their versatility and potential in various tasks within 3D domains, expanding their utility beyond conventional text-based applications.
These tasks encompass a broad spectrum, including answering questions about 3D scenes~\citep{hong2023_3D_LLM, dwedari2023generating}, planning and generating motion in 3D environments~\citep{hong2023_3D_LLM,zhang2023building,lv2024gpt4motion}, reconstructing scenes~\citep{igllm,avetisyan2024scenescript,raw}, synthesizing 3D scenes from textual descriptions~\citep{sun2023_3D_GPT,yang2023holodeck,huscenecraft,composeanything}, editing procedural models (those generated by sets of rules or algorithms) using natural language instructions~\citep{kodnongbua23reparamCAD}, and learning multi-modal representations that bridge text and 3D domains~\citep{xue2023ulip, hong2023_3D_LLM}.

\begin{figure*}[t]
\centering

\makebox[\linewidth][c]{%
\includegraphics[width=1.015\linewidth]{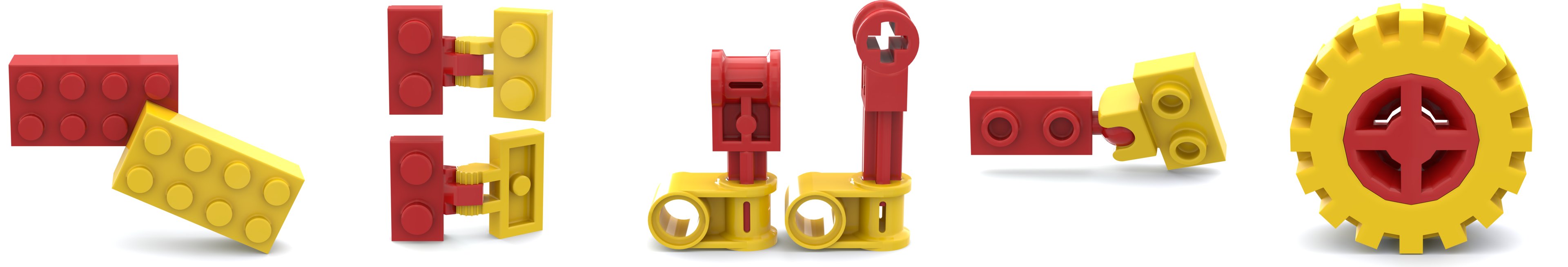}%
}
\begin{subfigure}[b]{.19\linewidth}
\centering
\caption{\textbf{Stud}}
\label{fig:stud}
\end{subfigure}
\hfill
\begin{subfigure}[b]{.196\linewidth}
\centering
\caption{\textbf{Hinge}}
\label{fig:hinge}
\end{subfigure}
\hfill
\begin{subfigure}[b]{.196\linewidth}
\centering
\caption{\textbf{Axle}}
\label{fig:axle}
\end{subfigure}
\hfill
\begin{subfigure}[b]{.196\linewidth}
\centering
\caption{\textbf{Ball}}
\label{fig:ball}
\end{subfigure}
\hfill
\begin{subfigure}[b]{.19\linewidth}
\centering
\caption{\textbf{Fixed}}
\label{fig:fixed}
\end{subfigure}

\caption{\textbf{Connectivity Semantics.}
We broadly model five types of connectivity between bricks.
Stud (\cref{fig:stud}) connections, after defining which stud connects to which hole, have at most one degree of freedom.
Hinge (\cref{fig:hinge}) connections have a degree of rotational freedom, and often the ability to be flipped (binary).
Axle (\cref{fig:axle}) connections inherit the same freedom as hinges, but can also be offset along their principal axis.
Ball (\cref{fig:ball}) connections have three degrees of rotational freedom.
Fixed (\cref{fig:fixed}) connections have no degrees of freedom.
}
\label{fig:connectivity}
\end{figure*}

\section{Representation}\label{sec:representation}

\subsection{Preliminaries}\label{ssec:ldraw}
We build on the LDraw~\cite{ldraw} ecosystem, a community-maintained standard and library that is ubiquitous for digital LEGO modeling.
The part library consists of over 24,000 CAD replicas, covering most real-world pieces.
Structure is represented in plain text as a set of references to these pre-defined parts, with attributes of instance color, 3D position, rotation matrix, and part type.
Positional units are expressed in LDU (LDraw Unit), where 1 LDU $\approx$ 0.4 mm and the diameter of a stud is 6 LDU.

Prior work has employed simplified representations.
Most generative-assembly approaches directly predict voxel occupancy then decompose it into grid-aligned brick positions~\cite{brecs}.
Some have also used graphs for generation~\cite{thompson}, but model only discrete structures.
BrickGPT~\cite{Pun_2025_ICCV}, the only existing text-serialized approach we are aware of, uses eight brick types and restricts placement to a discrete 20$\times$20$\times$20 grid.
In contrast, \citet{break_and_make,learning_to_buic} -- that do not target a generative problem -- make use of a broader set of parts.
They use the LDCad~\cite{ldcad} snap system to define connection sites and task an agent to visually select positions on pieces that should connect.
While much more general than a grid-based approach, the system lacks broad connector coverage.

\subsection{Connectivity}\label{ssec:connectivity}
In this section, we introduce our connector semantics.
We categorize connections between bricks into five families:

\noindent\textbf{Stud.} The most common type of connection is that between a stud and a hole.
As visualized in \cref{fig:stud}, once connected, at most one degree of rotational freedom is possible.
After keeping track of which stud is attached to which hole, only one rotational yaw parameter must be stored to fully represent the SE(3) transformation between two parts, greatly reducing dimensionality.
It is also true that two or more stud connections between parts determine rotation exactly, but we concern ourselves only with single-contact connections.

Within stud-type connectors, we subdivide into standard stud, open stud, hole, tube, and post.
Any stud type may pair with holes or tubes, but only open studs may connect to the posts found on the underside of some parts.
While a tube may also fit between four studs, the connection is non-standard and we do not model it.
We also do not model ``illegal'' connections (see \citet{stressing_elements} for discussion).

\noindent\textbf{Hinge.} In addition to the one degree of rotational freedom represented in stud-type connections, hinges must also encode a boolean ``flip.''
This is akin to separating a pair, rotating one 180 degrees in the axial, and re-connecting (\cref{fig:hinge}).
Hinges also have a broader range of in/on subtype pairs, which we model separately.

\noindent\textbf{Axle.} Axle connections extend hinges in that they involve one degree of rotation and a flip but also require parametrizing the ``slide'' along the axis (\cref{fig:axle}).
Within the category, we separately model pins, axles, pin sockets, axle sockets, bars, and clips.
Pins pair only with pin sockets, but axles can pair with both pin and axle sockets.
Clips connect with bars.

\noindent\textbf{Ball.} Ball-type connections similarly generalize hinges, but instead of a binary 180-degree ``flip,'' have two additional degrees of rotational freedom (\cref{fig:ball}).
We subtype connections into ``towball'' and socket, and ``technic'' and socket.

\noindent\textbf{Fixed.} Finally, some connections are ``fixed'' and allow for no meaningful freedom between the two parts in a pair, so knowing which connector is paired with which is sufficient to represent the spatial relationship.
One such example is the attachment of a wheel hub to a tire (\cref{fig:fixed}).

\subsection{Connector Annotation}\label{ssec:annotation}
To annotate the LDraw part library with our connector taxonomy, we perform a mixture of procedural annotation and manual authoring.
Bricks in the LDraw system are generally and conveniently defined by varying levels of subcomponents.
By identifying a ``stud.dat'' in the primitive hierarchy after verifying the scale in the composed rotation matrix, a precise connector position can be inferred.
While we reviewed all annotations, many edge cases exist, and not all connector sites unambiguously represent one type or another.

During stud-site filtering, we performed collision checks to determine whether a piece could indeed be connected at a given site.
However, we found collision estimation to be somewhat uniquely difficult in this setting.
First, even assuming perfect LDraw modeling, real parts can only connect to one another through some degree of stress-based plastic deformation.
This means that, even in a pair of parts that are well-connected, some level of collision is a given.
In traditional collision estimation, the amount of overlap between two meshes can be computed.
However, brick connections are very non-convex, and if a tube tightly connected to a stud collides with it on all sides, the depenetration solution is to remove the stud from the tube.
Additionally, the parts are largely non-watertight, so methods such as VHACD~\cite{vhacd} cannot be applied to decompose them into convex colliders.
To resolve these issues, we devise a multi-stage pipeline to attempt to render the part library watertight.
Following this process, we applied a modified version of PFPOffset~\cite{pfpoffset} to inset all part-mesh faces by 0.25 LDU (0.1 mm).
These meshes are then used for standard collision detection in both filtering and, later, generation evaluation.

\subsection{Graph}\label{ssec:graph}
\begin{figure}[t]
\centering

\begin{subfigure}[t]{.48\linewidth}
\centering
\includegraphics[width=\linewidth]{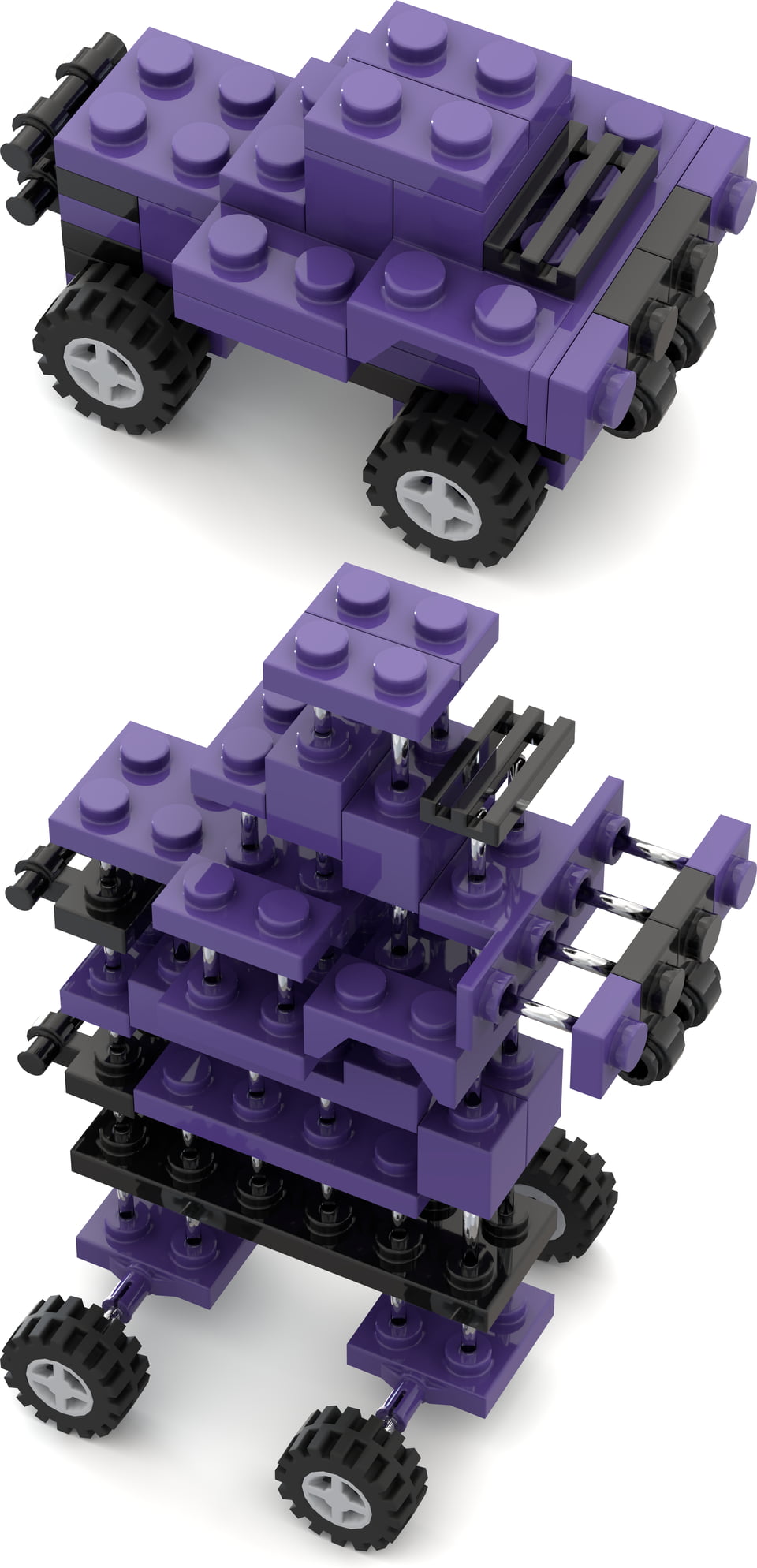}
\par\vspace{-0.1cm}
\caption{\textbf{Object}}
\label{fig:truck}
\end{subfigure}
\hfill
\begin{subfigure}[t]{.48\linewidth}
\centering
\includegraphics[width=\linewidth]{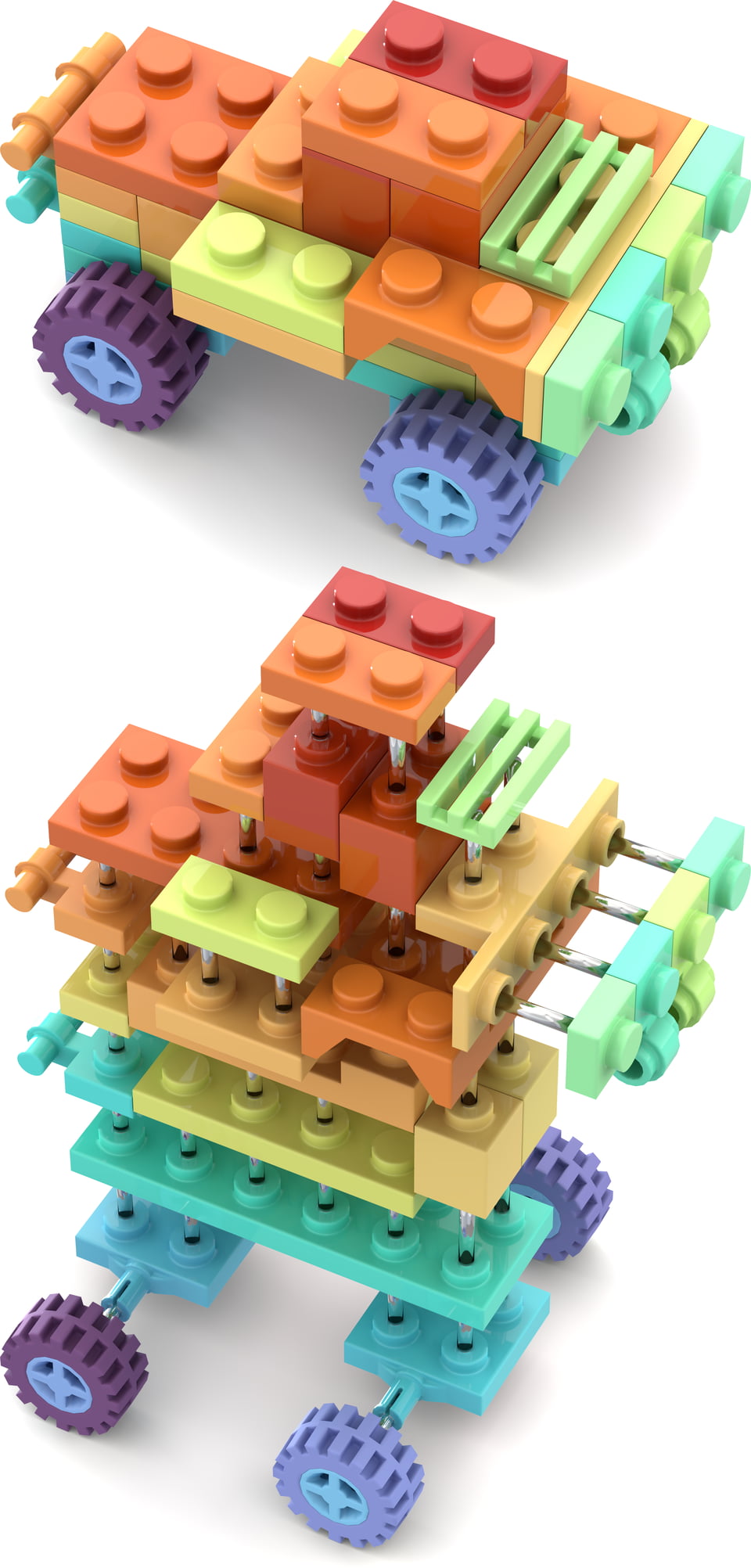}
\par\vspace{-0.1cm}
\caption{\textbf{Path-Colored}}
\label{fig:path-truck}
\end{subfigure}\\
\begin{subfigure}[b]{\linewidth}
\begingroup
\setlength{\tabcolsep}{0pt}
\renewcommand{\arraystretch}{1.00}
\newcommand{\SeqBracketPad}{0.3ex}
\newcommand{\SeqBracketSinglePad}{0.08ex}
\newcommand{\SeqBracketHeightScale}{0.86}
\newcommand{\SeqBracketLeft}{-0.2ex}
\newcommand{\SeqBracketWidth}{0.72ex}
\newcommand{\SeqDotsSep}{0.50em}
\newcommand{\SeqDotsScale}{1.25}
\newcommand{\SeqItem}[2]{{\ttfamily\strut\textcolor[HTML]{#1}{\CIRCLE}\ #2}}
\newcommand{\SeqDots}{{\ttfamily\strut$\vcenter{\hbox{\scalebox{\SeqDotsScale}{$\cdot\hspace{\SeqDotsSep}\cdot\hspace{\SeqDotsSep}\cdot$}}}$}}
\newcommand{\SeqBracketRows}[3][\SeqBracketPad]{%
\path
([xshift=\SeqBracketLeft,yshift=#1]m-#2-1.north west) coordinate (SeqTopRaw)
([xshift=\SeqBracketLeft,yshift=-#1]m-#3-1.south west) coordinate (SeqBottomRaw)
($(SeqTopRaw)!0.5!(SeqBottomRaw)$) coordinate (SeqMid)
($(SeqMid)!\SeqBracketHeightScale!(SeqTopRaw)$) coordinate (SeqTop)
($(SeqMid)!\SeqBracketHeightScale!(SeqBottomRaw)$) coordinate (SeqBottom)
([xshift=\SeqBracketWidth]SeqTop) coordinate (SeqTopRight)
([xshift=\SeqBracketWidth]SeqBottom) coordinate (SeqBottomRight);
\draw[overlay,line width=0.15ex,line cap=round]
 (SeqTop) -- (SeqBottom);
\draw[overlay,line width=0.15ex,line cap=round]
 (SeqTop) -- (SeqTopRight);
\draw[overlay,line width=0.15ex,line cap=round]
 (SeqBottom) -- (SeqBottomRight);
}
\begin{tikzpicture}[baseline=(m-1-1.base)]
\matrix (m) [
  matrix of nodes,
  nodes={anchor=west,inner sep=0pt,outer sep=0pt},
  row sep=0.30ex,
  column sep=0pt
] {
\SeqItem{E47773}{a plate 1x2 | purple} \\
\SeqItem{AB5142}{b brick 1x2 | purple} \\
\SeqItem{E47773}{a stud hole b stud b 90} \\
\SeqItem{f6AD73}{c brick 1x2 | purple} \\
\SeqItem{AB5142}{a stud hole a stud b 90} \\
\SeqDots \\
\SeqItem{96D9F4}{z plate 2x2 wheels holder | purple} \\
\SeqItem{86E7CD}{w stud hole d stud b 90} \\
\SeqDots \\
\SeqItem{8DA8E5}{ac wheel rim 6.4x8 | light grey} \\
\SeqItem{96D9F4}{z axle bar b clip a flip 270 0} \\
\SeqDots \\
\SeqItem{6C4E7A}{ah tyre 6 50x8 | black} \\
\SeqItem{8DA8E5}{ac fixed e in b a} \\
};
\SeqBracketRows[\SeqBracketSinglePad]{1}{1}
\SeqBracketRows{2}{3}
\SeqBracketRows{4}{5}
\SeqBracketRows{7}{8}
\SeqBracketRows{10}{11}
\SeqBracketRows{13}{14}
\end{tikzpicture}
\endgroup
\par\vspace{-0.1cm}
\caption{\textbf{Build Sequence}}
\label{fig:path}
\end{subfigure}
\par\vspace{-0.1cm}
\caption{\textbf{Graph Visualization.}
After encoding relative transformations between parts into their connectivity, we arrive at connected graphs.
From these graphs, we can sample iterative build instructions (spanning trees), that begin at a root part, add another part, define an edge that connects that part with the existing structure, and on.
For example, see the dark red piece at the top of the render in \cref{fig:path-truck}.
This corresponds to part 0 in the graph, the root node.
From that, it has two neighbors, both \texttt{brick 1x2}, which are added to the structure.
Each bracketed item corresponds to a discrete placement ``action.''
}
\vspace{-0.2cm}
\label{fig:graph}
\end{figure}

Given an unordered set of part instances, we follow the above matching semantics to pair connectors and form edges.
As designed, each edge defines the full local SE(3) transform between two paired parts.
Together, the edges form a graph of connectivity.
As each edge is sufficient, the full structure can be compactly and interpretably represented by a spanning tree of the graph.
See \cref{fig:truck} for a visualization of a parsed object and \cref{fig:path-truck} for a view colored by step order.

Starting with an arbitrary root, we sample a sequence of build steps that define the part to add and specify how it is to be connected onto the existing structure.
These build steps define a program that, when executed, produces a set of parts with 6-DoF pose.
During serialization, we round rotational parameters to the nearest degree and the slide scalar to the nearest LDU.
See \cref{fig:path} for a sample build sequence.

\section{Dataset}\label{sec:dataset}
\begin{figure*}[t]
\centering
\begin{subfigure}[b]{.2465\linewidth}
\centering
\includegraphics[width=\linewidth]{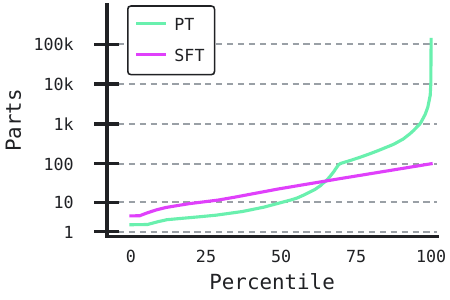}
\caption{}
\label{fig:bpo}
\end{subfigure}
\hfill
\begin{subfigure}[b]{.2465\linewidth}
\centering
\includegraphics[width=\linewidth]{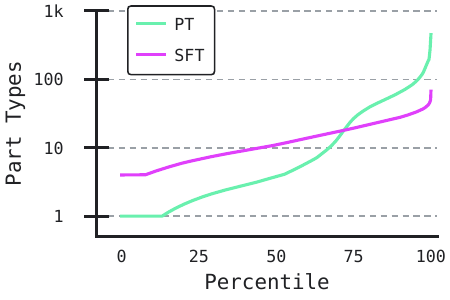}
\caption{}
\label{fig:uppo}
\end{subfigure}
\hfill
\begin{subfigure}[b]{.2465\linewidth}
\centering
\includegraphics[width=\linewidth]{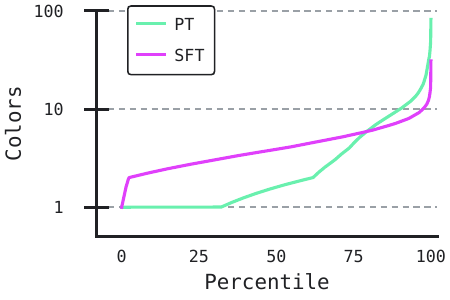}
\caption{}
\label{fig:ucpo}
\end{subfigure}
\hfill
\begin{subfigure}[b]{.2465\linewidth}
\centering
\includegraphics[width=\linewidth]{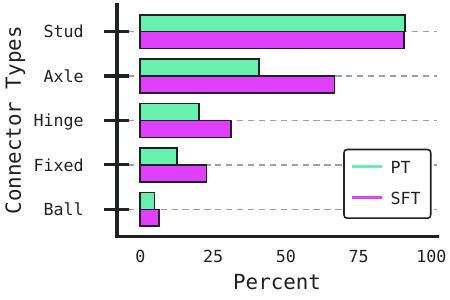}
\caption{}
\label{fig:ctpo}
\end{subfigure}

\caption{\textbf{Dataset Plots.}
We compute statistics over both BrickNet subsets.
While SFT samples are capped at 100 parts, the BrickNet-PT set is long-tailed and can include thousands of parts (\cref{fig:bpo}).
The number of unique parts (\cref{fig:uppo}) and colors (\cref{fig:ucpo}) per object are also similarly tailed.
We additionally compute the proportion of samples containing an instance of each connection-type class (\cref{fig:ctpo}).
}
\label{fig:dataset}
\end{figure*}

\begin{figure}[t]
\centering

\begin{subfigure}[b]{\linewidth}
\centering
\includegraphics[width=\linewidth]{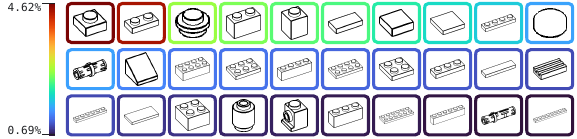}
\caption{Part Occurrence}
\label{fig:po}
\end{subfigure}
\vspace{0.2cm}
\begin{subfigure}[b]{\linewidth}
\centering
\includegraphics[width=\linewidth]{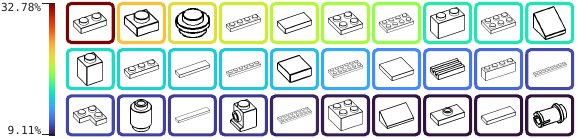}
\caption{Part Occurrence by Sample}
\label{fig:pobs}
\end{subfigure}

\caption{\textbf{Part Frequency.}
We compute relative part frequency (\cref{fig:po}) and the proportion of samples each part occurs in (\cref{fig:pobs}).
}
\label{fig:heatcard}
\end{figure}

\begin{table}[t]
\caption{\textbf{Dataset Statistics}.
We present BrickNet, a large-scale dataset of brick structures.
In contrast to the voxelized \hbox{BrickGPT}~\cite{Pun_2025_ICCV}, our samples include thousands of distinct part types.
Relative to the Official Model Repository (OMR)~\cite{ldraw} data used in \citet{break_and_make}, our set is much broader.
}
\label{table:dataset}
\centering
{
\setlength{\tabcolsep}{3.6923pt}
\begin{tabular}{lrrccc}
\toprule
Dataset     & Samples      & Parts & Color & Captions & Real \\
\midrule
BrickGPT~\cite{Pun_2025_ICCV} & 28{,}259 & 8           & \xmark & \cmark & \xmark \\
OMR~\cite{break_and_make} & 1{,}814 & 5{,}005 & \cmark & \xmark & \cmark \\
BrickNet-PT & 320{,}808      & 9{,}743    & \cmark & \xmark & \cmark \\
BrickNet-SFT & 67{,}185       & 6{,}457     & \cmark & \cmark & \cmark \\
\bottomrule
\end{tabular}
}
\end{table}

We collect a large-scale dataset of publicly available LDraw-format structures online.
This includes 40,549,969 instances of 9,743 unique parts across 320,808 samples.

We define two overlapping sets from this, BrickNet-PT (pretraining) and BrickNet-SFT (fine-tuning).
BrickNet-SFT contains 67,185 samples, with 1,774,387 cumulative instances, between four and 100 parts that fulfill part-color and part-type diversity criteria.
Each sample was additionally required to have zero collisions across the assembled structure.
We render each sample from eight views each and generate captions using Gemini 2.5~\citep{comanici2025gemini25pushingfrontier}.

We hold out 512 additional samples meeting the same criteria for evaluation.
Evaluation samples are drawn from source files containing a single object.

The remaining data is considerably more long-tailed.
We include it with the SFT train samples and label it BrickNet-PT.
It contains more than an order of magnitude more part instances, but often in the form of very large structures.
See \cref{table:dataset,fig:dataset,fig:heatcard} for dataset statistics.

\section{Experiments}\label{sec:experiments}

\subsection{Unconditional Generation}\label{ssec:unconditional}
\begin{table}[t]
\caption{\textbf{Unconditional Generation}.
We sample $2^{16}$ full-length sequences (100 parts) from each of our models.
We evaluate validity of each build prefix by whether it can be, first, realized as a connected structure, and second, be free of placement collisions.
The numbers reported represent the average number of successful build steps until an invalidating step.
We report results for both nucleus sampling (NS) and ancestral sampling (AS).
}
\label{tab:unconditional}
\centering
\begin{tabular}{@{}l cc cc@{}}
\toprule
& \multicolumn{2}{c}{Connectivity (AS / NS)} & \multicolumn{2}{c}{Collision (AS / NS)} \\
\cmidrule(lr){2-3} \cmidrule(lr){4-5}
Size & Graph & Pose & Graph & Pose \\
\midrule
0.6b & 45.0 / 94.1 & 13.6 / 31.8 & 11.0 / 16.0 & 8.1 / 14.5 \\
1.7b & 53.7 / 95.1 & 15.1 / 35.5 & 11.7 / 16.6 & 8.8 / 16.1 \\
4b   & 49.9 / 96.9 & 18.3 / 45.1 & 12.2 / 18.0 & 10.6 / 20.3 \\
8b   & 56.6 / 97.0 & 17.4 / 44.9 & 12.7 / 18.7 & 10.2 / 20.1 \\
14b  & 55.2 / 96.9 & 20.3 / 49.9 & 13.0 / 19.1 & 11.6 / 22.4 \\
\bottomrule
\end{tabular}
\end{table}

\begin{figure}[t]
\centering
\includegraphics[width=\linewidth]{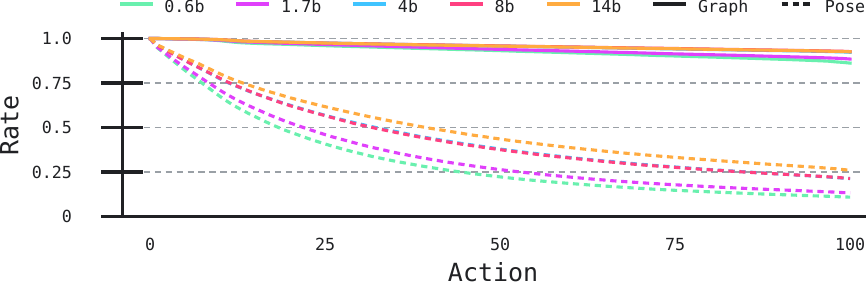}
\caption{\textbf{Connectivity Survival.}
On the nucleus-sampled sequences produced from our unconditional models, we compute the proportion of generations which ``survived'' at least $k$ placement actions before an unparseable or unsupported action was sampled.
}
\label{fig:survival}
\end{figure}

\begin{figure*}[t]
\centering
\begin{subfigure}[t]{\linewidth}
  \centering
  \includegraphics[width=\linewidth]{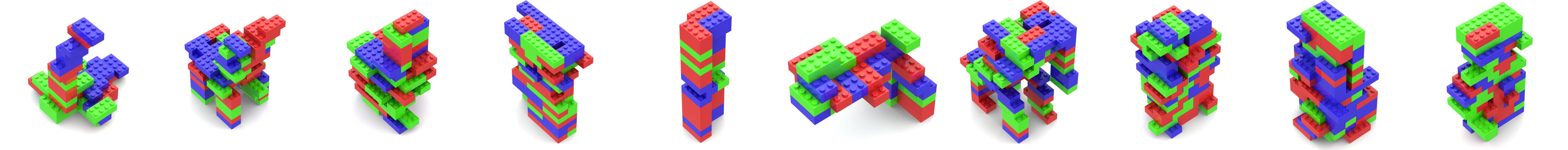}
  \par\vspace{-0.2cm}
  \caption{BrECS~\citep{brecs}}
  \label{fig:unconditional_mosaic_brecs}
\end{subfigure}
\par
\begin{subfigure}[t]{\linewidth}
  \centering
  \includegraphics[width=\linewidth]{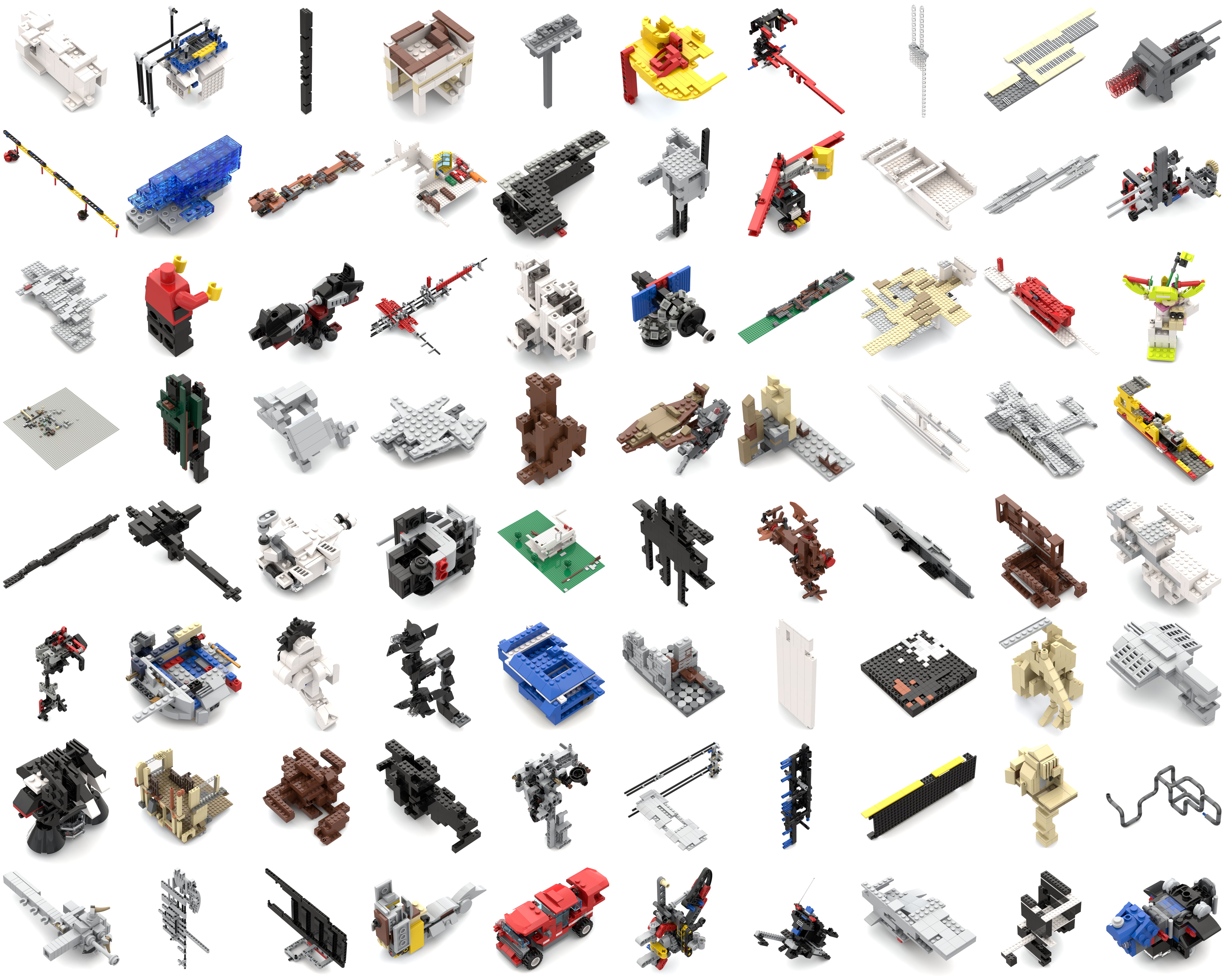}
  \par\vspace{-0.2cm}
  \caption{Ours}
  \label{fig:unconditional_mosaic_ours}
\end{subfigure}
\par\vspace{-0.2cm}
\caption{\textbf{Unconditional Samples.}
Random samples drawn from either BrECS~\citep{brecs} (\cref{fig:unconditional_mosaic_brecs}) or our model (\cref{fig:unconditional_mosaic_ours}).
\twemoji{mag_right} Zoom in for details.
}
\label{fig:unconditional_mosaic}
\vspace{-0.2cm}
\end{figure*}

Our first experiment concerns unconditional sequence generation.
Given a beginning-of-sequence token, the task is to produce a physically valid, collision-free structure.
For this, we use BrickNet-PT and sample build sequences of up to 100 pieces each.
We sample them in proportion to the square root of the number of pieces in a given structure.
While the internet-curated objects themselves exhibit numerous part--part collisions, we perform collision detection at path-sampling time to ensure training paths are valid.

We serialize the sampled sequences in both our proposed graph-backed form described in \cref{ssec:graph} and a naive non-graph pose-based form of part type, color, x/y/z, and yaw/pitch/roll.
Using these sequences, we finetune the 0.6b-, 1.7b-, 4b-, 8b-, and 14b-parameter Qwen 3~\cite{yang2025qwen3technicalreport} instruct models.
We cap sequence length at 4,096 tokens and train using only a standard next-token-prediction cross-entropy loss:
\begin{equation}\label{eq:next-token}
p(x) = \prod_{i=1}^n p\left(s_i|s_1,\ldots s_{i-1}\right)
\end{equation}

After training, we sample $2^{16}$ build sequences from each model at full temperature, suppressing the EOS token to force full-length generation.
With true full-temperature ancestral sampling (AS), we observe that the majority of generations for both the pose and graph models are invalidated by the sampling of garbage tokens.
At a sequence length of 4096, if only 0.1\% of next-token probability mass lies on invalid completions, the chance of sampling a valid sequence is less than 1.7\%.
We find that nucleus sampling (NS) with a top-\textit{k} of 20 and top-\textit{p} of 0.95 alleviates this, with average connectivity validity doubling (\cref{tab:unconditional}), which we adopt for subsequent experiments.

Following the change in sampling, the graph-backed approach achieves an average parseability / connectivity validity of 94 or more placement actions, while the direct pose approach does not reach 50.
To better understand this, we plot survival in \cref{fig:survival} and observe steep validity falloff.
In contrast, the graph-backed parametrization does not exhibit this steep falloff, as connectivity is encoded directly into the representation.
However, when collision avoidance is included, the approaches are comparable, with approximately 20 steps (\cref{tab:unconditional}).
Across metrics the 0.6b model performs worst, but there appear to be diminishing returns in how greater size translate to higher validity.
We show sample generations from the graph-backed 14b in \cref{fig:unconditional_mosaic} against those of BrECS~\citep{brecs}, a recent generative model supervised with ModelNet-40~\citep{modelnet} voxel structures.

\subsection{Text-Conditioned Generation}\label{ssec:conditional}
\providecommand{\roundmetric}[1]{\pgfmathprintnumber[fixed,fixed zerofill,precision=3]{#1}}

\begin{table}[t]
\caption{\textbf{Text-Conditioned Generation}.
Generation results on our 512-sample evaluation set. $P_\text{inv}$ represents the proportion of invalid placements.
VQAScore~\citep{vqascore}, PE~\cite{pe}, and SigLIP 2~\citep{siglip2} are measures of image--text similarity computed on object renders.
}
\label{table:conditional}
  \centering
  {
  \setlength{\tabcolsep}{4pt}
  \begin{tabular}{l c c c c }
    \toprule
    Model &  $P_\text{inv}$ & VQAScore & PE & SigLIP 2 \\
    \midrule
    BrickGPT~\citep{Pun_2025_ICCV} & 0.063 & \roundmetric{0.04964784097091979} & \roundmetric{0.15687567935674450} & \roundmetric{0.05195672460897249} \\
    \midrule
    0.6b  & 0.256 & \roundmetric{0.55708194033877589} & \roundmetric{0.27886060602031648} & \roundmetric{0.60347736479889136} \\
    1.7b  & 0.260 & \roundmetric{0.59318309911395772} & \roundmetric{0.28201144194463268} & \roundmetric{0.63094044892932288} \\
    4b    & 0.239 & \roundmetric{0.61480620496149641} & \roundmetric{0.28332354206941091} & \roundmetric{0.63897086286669946} \\
    8b    & 0.233 & \roundmetric{0.60835895728450851} & \roundmetric{0.28387912639300339} & \roundmetric{0.64687048157122717} \\
    14b   & 0.231 & \roundmetric{0.60203790253581246} & \roundmetric{0.28269848425406963} & \roundmetric{0.62462975183007075} \\
    \bottomrule
  \end{tabular}
}
\end{table}

\begin{figure*}[t]
\centering
\begingroup
\newcommand{\PairGridWithCards}[3]{%
  \begin{tikzpicture}
    \node[inner sep=0pt, outer sep=0pt, anchor=south west] (#1) at (0,0) {%
      \includegraphics[width=\linewidth]{#2}%
    };
    \path
      ($(#1.north west)!0.5!(#1.south west)$) coordinate (#1MidL)
      ($(#1.north east)!0.5!(#1.south east)$) coordinate (#1MidR);
    \foreach \k in {1,...,8}{%
      \pgfmathsetmacro{\x}{(\k-1)/8 + 0.008}%
      \pgfmathtruncatemacro{\labelnum}{#3 + \k - 1}%
      \node[
        cvprPill,
        rounded corners=1.8pt,
        anchor=west,
        inner xsep=1.8pt,
        inner ysep=0.9pt,
        font=\scriptsize\bfseries
      ] at ($(#1MidL)!\x!(#1MidR)$) {\labelnum};
    }%
  \end{tikzpicture}%
}

\PairGridWithCards{uppergrid}{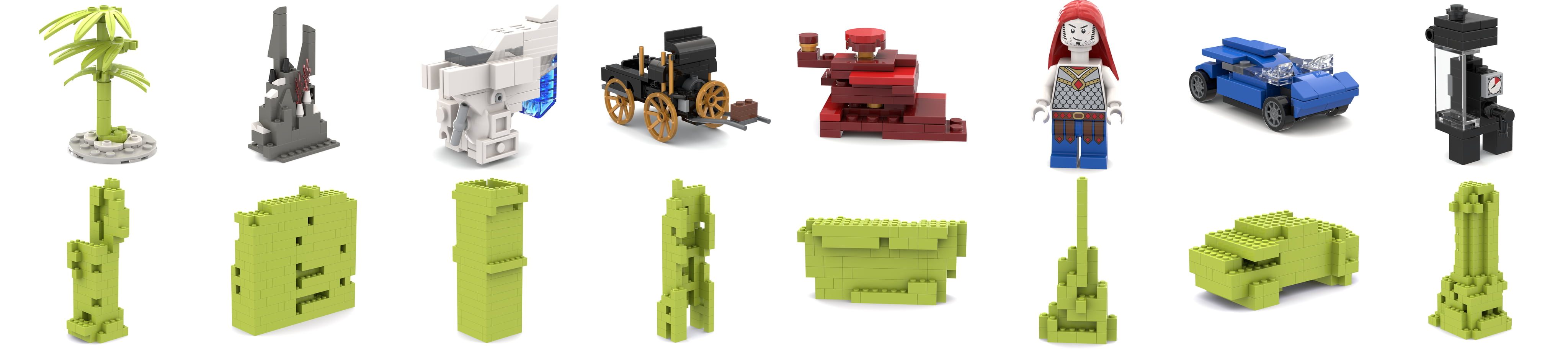}{1}
\par\vspace{0.3cm}
\PairGridWithCards{lowergrid}{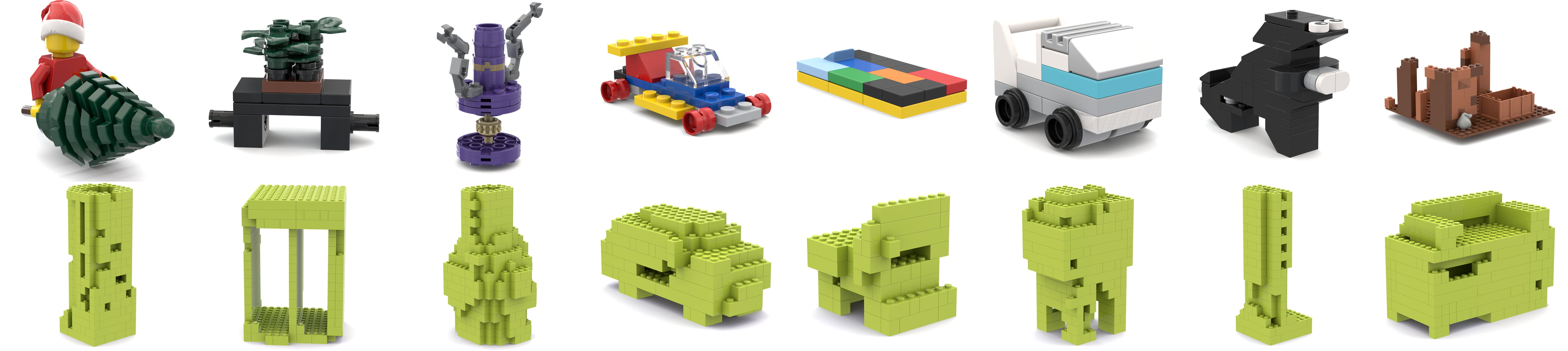}{9}

\vspace{0.2cm}
\noindent\makebox[\linewidth][l]{%
\tikz[baseline=(cap.base)]\node[
  cvprPill,
  rounded corners=1.8pt,
  inner xsep=2.0pt,
  inner ysep=1.5pt,
  outer sep=0pt,
  anchor=west
] (cap) at (0,0) {%
\begin{minipage}[t]{\linewidth}
\fontsize{10}{10}\selectfont
\renewcommand{\baselinestretch}{1}\selectfont
\setlength{\parindent}{0pt}
\setlength{\parskip}{0pt}
\setlength{\baselineskip}{10pt}
\setlength{\lineskip}{0pt}
\setlength{\lineskiplimit}{0pt}
\leftskip=0pt\rightskip=0pt plus 0.12em\parfillskip=0pt plus 1fil
\def\Pid#1{\textbf{\textsf{#1}}\nobreak\hspace{0.35em}\ignorespaces}%
\noindent\ignorespaces
\Pid{1} This is a LEGO model of a stylized, light-green bamboo stalk with leaves, constructed from cylindrical bricks and leaf elements, mounted on a circular white base.
\Pid{2} This LEGO model depicts a small, gray stone shrine or altar, featuring an archway topped with two torches, a central pedestal holding a red gem, and various accessories including a green plant, a white urn, and a black fixture.
\Pid{3} This is a LEGO model of a white and grey handheld device, possibly a scanner or medical tool, featuring a blue transparent element and a silver cylindrical piece at its front.
\Pid{4} This is a LEGO model of a traditional black and gold rickshaw, featuring two large spoked wheels, a covered passenger seat with beige upholstery, and long pulling handles at the front.
\Pid{5} This is a LEGO model of a rectangular, dark red gift box, decorated with a lighter red ribbon and bow on top, and featuring a black winding key on its side, indicating it is a music box.
\Pid{6} This LEGO minifigure features vibrant red hair, a white torso printed with a silver scale-mail vest, and solid blue legs.
\Pid{7} This is a blue LEGO sports car model, featuring a prominent rear spoiler, grey wheels, and transparent windows, shown from multiple angles.
\Pid{8} This is a detailed LEGO model of a black espresso machine, featuring a transparent bean hopper on top, a side-mounted water reservoir, a front-facing portafilter and steam wand, and a pressure gauge.
\Pid{9} This LEGO model features a minifigure in a red torso and Santa hat, holding a large, stylized green Christmas tree made of stacked, angled pieces.
\Pid{10} This LEGO model is a rectangular patch of dark green grass, constructed from numerous plant stem pieces on a black plate with a brown base, featuring a black Technic axle protruding from one end for connection.
\Pid{11} This is a LEGO model of a large, cylindrical purple container or barrel, reinforced with tan bands and featuring various grey mechanical attachments, including handles, levers, and what appear to be nozzles or connectors.
\Pid{12} This is a simple, abstract LEGO vehicle constructed from primary-colored bricks---blue, red, yellow, and black---and equipped with four red wheels.
\Pid{13} This LEGO model is a colorful, low-profile bed constructed from a yellow base and a top layer of multicolored rectangular bricks, featuring a light blue wedge piece at one end to serve as a headboard.
\Pid{14} This is a 3D digital model of a stylized, blocky LEGO vehicle, resembling a futuristic truck or armored van, shown from multiple angles to display its gray, white, and light blue construction with prominent black wheels.
\Pid{15} This is a LEGO model of a black squirrel, constructed with dark grey and light grey bricks, featuring a large bushy tail, prominent white eyes, and holding two small, round objects in its paws.
\Pid{16} This LEGO model depicts a dilapidated, single-room shelter or bunker with crumbling brown and tan walls, containing a simple bed, a rifle, a sack, and a small cache of money.\unskip
\end{minipage}%
};}%
\endgroup
\caption{\textbf{Text-Conditioned Samples.} Samples produced using prompts from the evaluation set.
Outputs are arranged as pairs with a prompt number.
Within pairs, our outputs are above and those of BrickGPT~\citep{Pun_2025_ICCV} are beneath.
Match the number to the full prompt below.}
\label{fig:conditional_mosaic}
\end{figure*}

Next, we finetune our graph-backed PT models on BrickNet-SFT for the task of text-conditioned structure generation.
We follow the same training objective as in the PT experiment.

We evaluate on 512 held-out test prompts and score renders of our generations perceptually against the text prompts using 1) VQAScore~\citep{vqascore} with Qwen2.5-VL 7b~\citep{qwen2.5-VL}; 2) Perception Encoder (PE)~\citep{pe}; and 3) SigLIP 2~\citep{siglip2}.
If a sequence is not parseable, we regenerate it, which was necessary to evaluate all model-size variants.
We compare against prior work BrickGPT~\citep{Pun_2025_ICCV} and provide samples for both in \cref{fig:conditional_mosaic}.

While we observed fairly consistent quantitative trends in the unconditional experiment across model sizes (\cref{table:conditional}), the pattern in perceptual quality is less clear.
The 0.6b model underperforms all others, but 1.7b outperforms 14b on one metric and 14b is the best only in avoiding invalid placements.
This suggests that model capacity may not be a key limiting factor holding back quantitative performance.
However, with the exception of the ordering between 0.6b and 1.7b, the proportion of invalid placements appears to trend with size, in line with findings from the unconditional stage.

We additionally evaluate model perplexity on the ground-truth held-out sequences in \cref{table:perplexity} for both the PT and SFT graph models, and for varying amounts of PT or SFT data.
In contrast to the perceptual evaluation, we find a consistent ranking as model and dataset size are increased.
We suggest that this may indicate misalignment between the training objective -- which is being correctly minimized -- and the sampling task.
We note also that perplexity is parametrization-variant and cannot be used to directly evaluate between pose and graph representations.
\begin{table}[t]
  \caption{\textbf{Perplexity.}
  We compute perplexity for each of the graph-backed model sizes on our held-out 512 samples.
  PT models are evaluated without a prompt.
  We ablate both the effect of our training stages (\cref{table:perplexity_training}) and the amount of PT/SFT training data (\cref{table:perplexity_dataset}).
  }
  \label{table:perplexity}
  \centering
  \begin{subtable}[t]{\linewidth}
    \caption{\textbf{Training-Stage Ablation}}
    \label{table:perplexity_training}
    \centering
    \setlength{\tabcolsep}{4pt}
    \begin{tabular}{l c c c}
      \toprule
      Model & PT (uncond.) & PT + SFT (cond.) & No-PT + SFT \\
      \midrule
      0.6b & 1.331 & 1.298 & 1.343 \\
      1.7b & 1.324 & 1.290 & 1.326 \\
      4b   & 1.307 & 1.274 & 1.311 \\
      8b   & 1.305 & 1.273 & 1.303 \\
      14b  & 1.300 & 1.266 & 1.298 \\
      \bottomrule
    \end{tabular}
  \end{subtable}
  \begin{subtable}[t]{\linewidth}
    \caption{\textbf{Dataset-Size Ablation (14b)}}
    \label{table:perplexity_dataset}
    \centering
    \setlength{\tabcolsep}{5pt}
    \begin{tabular}{l c c c c}
      \toprule
      PT$\backslash$SFT & Full & Half & Quarter & None \\
      \midrule
      Full    & 1.266 & 1.279 & 1.288 & 1.300 \\
      Half    & 1.273 & 1.284 & 1.296 & 1.318 \\
      Quarter & 1.276 & 1.292 & 1.305 & 1.361 \\
      \bottomrule
    \end{tabular}
  \end{subtable}
  \vspace{-0.15cm}
\end{table}

\section{Discussion and Limitations}\label{sec:limitations}
As the result of our large-scale object dataset and graph-backed parametrization, our model is able to learn to model sequences and connectivity fairly well.
However, it struggles to produce long sequences without introducing inter-part collisions.
Future work should consider both how inference-time approaches might be leveraged to guide decoding and how to improve the spatial understanding of the model itself.

While our graph representation is compact and fairly general, and may have broader application, it requires the model to have knowledge of the domain.
If not for the pretraining, the model would not know the positions of the connectors on the parts, and therefore would then not have the ability to use them.
In a closed setting such as this where the part vocabulary is static, this can be learned, but it poses challenges for generalization to new domains.

In order to maintain our focus on data and representation, our supervised fine-tuning evaluation was made deliberately straightforward.
There is likely strong potential for improvement through post-training techniques.

As a simplification, and due partially to computational constraints, we limited build-sequence length to 100 parts.
However, real-world sets carry no such limitation.
Future work should consider how scaling might be achieved.

While we demonstrated that unconditional pretraining improves model generalization and sample efficiency for the task of text-conditioned generation, we anticipate that the learned prior may be adapted to additional downstream tasks, including reconstruction or editing.
Additionally, our model, in autoregressively sampling full sequences, was effectively allowed an infinite library of parts.
Future work should also explore generation or assembly conditioned on a specified part set or under additional constraints.

\section{Conclusion}\label{sec:conclusion}
Our investigation represents the first attempt to model and autoregressively generate LEGO-brick structures using pieces with arbitrary connectivity.

To achieve this, we first proposed BrickNet, a large-scale dataset of human-designed digital LDraw structures consisting of millions of placed pieces.
Second, we introduced a low-dimensional parametrization of brick structure and demonstrated improved parsing validity across sequence lengths and model-size variants.
To make this possible, we annotated the LDraw part library with a typed connector system with precise positions.
Finally, in adapting our BrickNet-PT model to the task of text-conditioned generation, we demonstrated that our learned prior is adaptable.
We hope our dataset will enable broader task application.

\par{\small\noindent\textbf{Acknowledgments.}
This work was funded in part by the French government under management of Agence Nationale de la Recherche as part of the “France 2030" program, reference ANR-23-IACL-0008 (PR[AI]RIE-PSAI project) and the ANR project VideoPredict ANR-21-FAI1-0002-01. Cordelia Schmid would like to acknowledge the support by the Körber European Science Prize. We \hbox{are grateful to Michael Black for his encouragement and support.}
}
\vspace{-0.8cm}

{
    \small
    \bibliographystyle{ieeenat_fullname}
    \bibliography{main}
}

\end{document}